\documentclass[conference]{IEEEtran}
\IEEEoverridecommandlockouts
\usepackage{cite}
\usepackage{amsmath,amssymb,amsfonts}
\usepackage{algorithmic}
\usepackage{graphicx}
\usepackage{textcomp}
\usepackage{xcolor}
\def\BibTeX{{\rm B\kern-.05em{\sc i\kern-.025em b}\kern-.08em
    T\kern-.1667em\lower.7ex\hbox{E}\kern-.125emX}}
\begin{document}

\title{Distributed Evolution Strategies Using TPUs for Meta-Learning\\

%\thanks{Our research was supported with Cloud TPUs from Google’s TensorFlow Research Cloud (TFRC).}
}

\author{\IEEEauthorblockN{1\textsuperscript{st} Alex Sheng}
\IEEEauthorblockA{\textit{College of Arts and Science} \\
\textit{New York University}\\
New York, USA \\
alexsheng4@gmail.com}
\and
\IEEEauthorblockN{2\textsuperscript{nd} Jun Yi Derek He}
\IEEEauthorblockA{\textit{Academy of Science and Technology} \\
\textit{The Woodlands College Park High School}\\
The Woodlands, USA \\
derekhe99@yahoo.com}
}

\IEEEoverridecommandlockouts
\IEEEpubid{\makebox[\columnwidth]
{978-1-7281-2547-3/20/\$31.00~\copyright2020 IEEE \hfill} %--> insert the copyright option applicable from above.
\hspace{\columnsep}\makebox[\columnwidth]{ }}
% Add the following code after the \maketitle command:
\IEEEpubidadjcol

\maketitle

\begin{abstract}
Meta-learning traditionally relies on backpropagation through entire tasks to iteratively improve a model’s learning dynamics. However, this approach is computationally intractable when scaled to complex tasks. We propose a distributed evolutionary meta-learning strategy using Tensor Processing Units (TPUs) that is highly parallel and scalable to arbitrarily long tasks with no increase in memory cost. Using a Prototypical Network trained with evolution strategies on the Omniglot dataset, we achieved an accuracy of 98.4\text{\%} on a 5-shot classification problem. Our algorithm used as much as 40 times less memory than automatic differentiation to compute the gradient, with the resulting model achieving accuracy within 1.3\text{\%} of a backpropagation-trained equivalent (99.6\text{\%}). We observed better classification accuracy as high as 99.1\text{\%} with larger population configurations. We further experimentally validate the stability and performance of ES-ProtoNet across a variety of training conditions (varying population size, model size, number of workers, shot, way, ES hyperparameters, etc.). Our contributions are twofold: we provide the first assessment of evolutionary meta-learning in a supervised setting, and create a general framework for distributed evolution strategies on TPUs.
\end{abstract}


\section{Introduction}
Meta-Learning is a rapidly growing field in deep learning which allows for the creation of models that can modify their own learning dynamics [1, 2, 3]. One critical aspect of human intelligence, which meta-learning algorithms aim to emulate, involves being able to classify unseen objects after only a few instances of viewing it. This is the goal of few shot learning [4, 5, 6]. This paper assesses the viability of an evolutionary strategy meta-learning approach on supervised few-shot classification problems.

Matching Nets (MN) use neural networks augmented with an attention mechanism to embed an unseen data point given a support set and uses a weighted distance calculation to determine the closest known class label to the new test point [4]. This effectively combines beneficial traits of parametric learning and nonparametric learning to solve the few shot classification problem. 

A model proposed in 2017, the Prototypical Network (ProtoNet), is also geared towards solving few shot classification problems [7]. A neural network is trained to map its embedding of instances of various classes into a latent vector space, where the mean of the instances of each class is computed to determine the class “prototype”. A distance calculation is performed with the prototype and new embedded cases to determine how to classify those new cases.

Model-Agnostic Meta-Learning (MAML) is another key development. MAML attempts to meta-learn a model initialization that converges after a small number of gradient steps on a variety of tasks [8]. When the model is trained to adapt to new tasks, a gradient of each task’s loss with repect to the initial parameters can be computed to iteratively adapt the model initialization. The key advantage of MAML is that it inherits the convergence guarantees of gradient descent, introducing a helpful inductive bias for the meta-learner.

Early work in the evolutionary meta-learning area used genetic algorithms in an adversarial environment to evolve self-referential meta-learner agents [9]. Furthermore, the work provided a biological rationale for evolutionary meta-learning as a paradigm to evolve increasingly intelligent agents.

ES-MAML [10] introduced by Song et al. showed that evolutionary strategies could be used to replace the outer loop backpropagation step in MAML, removing the need for computing second-order gradients. ES-MAML showed promising results on meta-learning tasks in reinforcement learning, but few experiments (few-shot sine regression) were done on supervised learning tasks. In this paper, we extend the evolutionary meta-learning regime to the supervised learning by using evolution strategies to train ProtoNet.

\section{Algorithms}

\subsection{Few-shot Classification}

In few-shot learning, the goal is to train a meta-learner to quickly learn new tasks [1, 2]. A common objective is to learn to accurately classify a set of image classes given only  a few samples from each class [11]. This is typically done with 1 or 5 samples per class, referred to as 1-shot or 5-shot classification respectively.
Few-shot learning is broken down into episodes. In each episode, multiple classes are sampled. Typically, either 5 classes (5-way classification) or 20 classes (20-way classification) are used. 

In $N$-way $K$-shot classification, $N$ classes are sampled per episode to create a $N$-way classification task. $K$ labeled samples from each class are given to a meta-learner to learn the task. These samples used to learn the task are referred to as the support set. The conditioned meta-learner is then evaluated on a separate query set (multiple unlabeled, unseen samples from the current episode’s classes), and the resulting average loss is used to update the meta-learner parameters.

Over many episodes, the meta-learner becomes able to quickly learn new tasks with little data. In order to evaluate the meta-learner’s ability to learn new tasks, validation episodes sample unseen classes from a validation set of classes, and the performance of the meta-learner on the query sets of these classes after conditioning on the corresponding support sets is averaged to determine the validation set loss and accuracy. The same procedure is carried out for test episodes, using a separate test set of classes, which are only used once at the end of training.

\subsection{Prototypical Networks}

Prototypical Networks use a neural network $f$ parameterized by $\varphi$ that maps an input $x$ of $M$ dimensions into an $N$-dimensional latent space [7]. In an episode’s support set, each instance is mapped into latent space, and the mean of the latent vectors of all instances in each class $k$ is deemed the prototype $c_k$ for that class. The number of classes in these sets is denoted by $n$.

The similarity D between a query instance and each class prototype is found using a similarity metric, either negative squared Euclidean distance [12] or cosine similarity [13], and a softmax function is applied to the query-to-prototypes similarity vectors to generate class probability distributions [14, 15]. Categorical cross entropy is applied to the resulting class probabilities to compute the loss [16].

\begin{equation}
D(x,y) = -(f_\varphi(x)-y)^2 \label{eq1}
\end{equation}

\begin{equation}
p_\varphi(f_\varphi(x)=k|x) = \frac{\exp{(D(x,c_k))}}{\sum_{i=0}^n\exp{(D(x,c_{k_i}))}} \label{eq2}
\end{equation}

\begin{equation}
J(\varphi) = -\log{p_\varphi{}(f_\varphi{}(x)=k|x)} \label{eq3}
\end{equation}

\subsection{Evolution Strategies}

Evolution strategies, originally proposed in the 1970s, is an optimization methodology that is inspired by biological evolution [9, 17, 18]. In ES, a population of candidate models is instantiated by sampling from a standard distribution in parameter space with mean $\mu$ and standard deviation $\sigma$. All candidate models are evaluated based on a fitness function and assigned a reward. Based on these rewards, $\mu$ and $\sigma$ are updated via an optimization rule. A new population is then instantiated using the updated mean and standard deviation, and the process repeats. In the variation of ES that we use in this paper, the initial mean $\mu$ is randomly initialized, the standard deviation $\sigma$ is a fixed scalar hyperparameter, and the fitness function is the negative of the few-shot cross-entropy loss. Our update rule performs gradient descent using the gradient of the expected loss computed via REINFORCE, in a very similar manner to Natural Evolution Strategies.

Natural Evolution Strategies (NES) incorporates information from every candidate solution to estimate the search gradient [19]. Given a probability distribution function $\pi(z,\theta)$ parameterized by $\theta$, the loss function J can be calculated by finding the expected value of the objective function F with respect to the model parameters z, which can be expressed below [20]:

\begin{equation}
J(\theta) = E_\theta{}[F(z)] = \int{}F(z)\pi(z,\theta)dz \label{eq4}
\end{equation}

The gradient of the loss function can be found by using the ‘log-likelihood trick’ [20], shown below:

\begin{equation}
\nabla_\theta{}J(\theta) = E_\theta{}[F(z)\nabla_\theta{}\log{\pi(z,\theta)}] \label{eq5}
\end{equation}

Given a population $z_i$ of $N$ candidate solutions [10], the gradient can be approximated as:

\begin{equation}
\nabla_\theta{}J(\theta) \approx \frac{1}{N}\sum_{i=0}^N F(z_i)\nabla_\theta{}\log{\pi(z_i,\theta)} \label{eq6}
\end{equation}

\subsection{Evolutionary Meta-learning Rationale}

The key advantage of meta-learning is the ability to incorporate the learning dynamics of the model itself into the training signal. In the case of few-shot learning, this is achieved by casting entire tasks as training samples for a meta-learner. This is traditionally done by storing activations and model states throughout a task, and then backpropagating through the entire task to compute meta-level gradients that account for the learning dynamics of the model over the course of the task. However, such a formulation of meta-learning requires storing backpropagation graphs that increase in size when scaling to longer tasks (the backpropagation graph will typically grow proportionally to task length), resulting in prohibitive memory costs when attempting meta-learning on complex tasks.

The learning dynamics of a meta-learner can also be interpreted as being similar to the hidden state dynamics of a recurrent neural network, with backpropagation through a task in meta-learning being analogous to backpropagation-through-time in RNN training. With such an interpretation, it would theoretically be possible to use forward-mode differentiation similar to Real-time Recurrent Learning (RTRL) [21] to do meta-learning with memory cost that is invariant with respect to task length. Unfortunately, such an approach would inherit RTRL’s prohibitive computational complexity with respect to model size. However, such a formulation successfully offloads the bulk of the memory cost of meta-learning from the backpropagation graph to the large jacobian matrices used in forward-mode differentiation (the largest jacobian is typically comprised of the gradients of the present-step hidden state w.r.t. the model parameters). The key advantage here would be the ability to compute meta-level gradients with memory cost that is invariant with respect to task length, although the inherent complexity and poor parallelizability of forward-mode differentiation makes it a poor candidate to meaningfully reduce the computational cost of meta-learning.

We propose an evolutionary meta-learning approach as a potential solution to the aforementioned problem. Specifically, meta-learning using evolution strategies would achieve backpropagation-free meta-learning (thus memory cost invariant with respect to task length) without the inherent computational complexity of forward-mode differentiation, while also being highly parallel. ES-based meta-learning would further offload the bulk of the memory cost of meta-learning from the large jacobians used in forward-mode differentiation to the population matrix used in ES. This would make meta-learning much more scalable and computationally tractable, as evolution strategies are naturally parallelizable along the population axis, and the population size in ES can be further adjusted to manage memory cost (although it is important to note that there is a tradeoff between memory cost and gradient accuracy when adjusting the population size in ES).

\subsection{Evolution vs. Analytical Methods for Meta-learning}

The dominant term $\Omega_{BP}$ in the memory cost of meta-learning by backpropagation is the sum $g$ of the memory sizes of all intermediate tensors computed in each inner-loop step multiplied by the number of inner-loop steps $l$ in a task, such that $\Omega_{BP} = g \times l$. Because of this linear scaling of memory cost with regard to task length, it is possible for the memory cost of backpropagation meta-learning to grow indefinitely as task size increases.

When meta-learning tasks exceed a certain number of steps in length, it is more memory-efficient to use forward-mode differentiation than to use backpropagation. This is true specifically when the memory size $\Omega_{BP}$ of the backpropagation graph is greater than the memory size $\Omega_{FM}$ of the Jacobian matrix formed by aggregating the gradients of the inner-loop model state $\psi \in \mathbb{R}^{D_\psi}$ w.r.t. the meta-learner parameters $\varphi \in \mathbb{R}^{D_\varphi}$. The “inner-loop model state” refers to a flattened vector of size $D_\psi$ containing information about the adapted model during a task. In MAML, for example, this model state would contain the task-adapted parameters $\theta’=\psi$ with the same size as the meta-learning parameters $\theta=\varphi$ such that $D_\psi = D_\varphi$, resulting in a forward-mode Jacobian of size $D_\psi \times D_\varphi = D_\varphi^2$. On the other hand, the model state in ProtoNet would contain the stored class prototypes $C$ from a task, such that the size of the model state is equal to the size $D^c$ of each class prototype multiplied by the number of classes $N$, resulting in a Jacobian of size $D_\psi \times D_\varphi = D_c \times N \times D_\varphi$.

The dominant term $\Omega_{FM}$ in the memory cost of forward-mode differentiation depends only on the number of meta-learner parameters $D_\varphi$ and the number of elements in the model state $D_\psi$. These scaling properties make forward-mode differentiation a suitable substitute for backpropagation (the memory cost of which also depends on task length) to reduce memory cost when performing meta-learning on long tasks. Because that both backpropagation and forward-mode differentiation are analytical differentiation methods, they will compute identical gradients and ultimately have the same convergence properties.

The dominant term $\Omega_{ES}$ in the memory cost of evolution strategies meta-learning is the size of the population matrix, equal to the product of the population size $n$ and the number of meta-learner parameters $D_\varphi$. Because that $P$ is a tuneable hyperparameter, it can be set smaller than $D_\psi$ to achieve a lower memory cost than forward-mode differentiation: $\Omega_{ES} < \Omega_{FM}$ given $P < D_\psi$.

Backpropagation is ideal for simple tasks with few inner-loop steps, as it is computationally cheap for short tasks and computes exact (analytical) gradients. Forward-mode differentiation also computes exact gradients, but is more memory-intensive than backpropagation on short tasks where $g \times l < D_\psi \times D_\varphi$ and less memory-intensive on long tasks where $g \times l_1 > D_\psi \times D_\varphi$. Evolution strategies, on the other hand, computes gradients numerically (therefore producing less exact gradients), but can be made less memory-intensive than forward-mode differentiation in almost any task by lowering the population size hyperparameter such that $P < D_\psi$. With this lower constant memory cost, evolution strategies becomes less memory-intensive than backpropagation at a lower threshold of task length than forward-mode differentiation: $g \times l_2 = P \times D_\varphi$ where $l_2 < l_1$ when $P < D_\psi$. Even in cases where $P > D_\psi$, evolution strategies may still be more feasible to implement due to being more parallel than backpropagation or forward-mode differentiation. Evolution strategies effectively allows for less memory cost by trading off gradient accuracy, potentially resulting in slower convergence and lower resulting model performance. However, we empirically show in this paper that the accuracy lost when using evolution strategies in place of analytical gradients is not substantial.

Backpropagation and forward-mode differentiation use the same set of forward and backward operations (backpropagation backpropagates through the entire task while forward-mode backpropagates through individual steps to update the Jacobian matrix) to compute gradients, resulting in similar time demands to train a meta-learner on a given task. Evolution strategies only requires a forward pass, resulting in about half as many operations as backpropagation. However, these operations are more expensive, as the forward propagation is performed with one input per candidate model in parallel. Thus, the exact time demands of evolution strategies depends heavily on parallelization and communication overhead.

\subsection{TPU Evolution Strategies}

A key advantage of parallel evolution algorithms is that a large number of independent fitness-level evaluations can be made in parallel across large clusters of processors [22]. Because of this, parallel evolution algorithms are most efficient when problems involve long fitness-level evaluations, low cross-worker communication overhead, and potential benefit when running a large number of evaluations in parallel. Due to these properties, parallel evolution strategies via TPUs are an ideal approach for meta-learning. Meta-learning entire tasks results in long fitness evaluations, TPU pod high-speed mesh networks mitigate the communication overhead of parallel ES, and high-dimensional optimization of deep meta-learners can benefit greatly from parallel evaluations for accurate gradient computation.

As a key contribution of this paper, we present the first-ever framework for distributed evolution strategies on TPUs. We use a similar algorithm to OpenAI’s distributed ES [23], with a modified update rule. In OpenAI-ES, the population is sampled using a fixed seed,  allowing for individual workers to re-sample the exact candidate model parameters of every other worker. Thus, it is possible to greatly reduce cross-worker communication overhead by only communicating final reward values. However, the re-sampling step in OpenAI-ES causes the per-worker computational cost to increase proportionally to the total number of workers, which results in poor efficiency when scaling to large population sizes. Our algorithm takes advantage of the TPU’s efficient communication topology to remove this constraint, by evaluating multiple candidates on each worker and computing gradient estimates separately without communicating rewards. The final gradient is then averaged across workers before the update step. Our formulation of distributed ES is especially suitable for TPUs, as it takes advantage of high communication bandwidth to cut out the resampling step, achieving low per-worker computational cost when scaling to arbitrarily large clusters. Specifically, we use the unique TPU AllReduce operation to efficiently average the gradient node across all cores.

\subsection{Evolution Strategy Prototypical Networks}

In our evolution strategy, we randomly initialize the mean model parameter vector $\mu$ of the neural network. We generate candidate solutions $x$ by adding displacement $\varepsilon$ using a Gaussian distribution centered on 0 with standard deviation $\sigma$.

\begin{equation}
x_i = \mu + \varepsilon \label{eq7}
\end{equation}

\begin{equation}
\varepsilon = N(0,\sigma) \label{eq8}
\end{equation}

Instead of weighting each candidate solution with the probability distribution function as in NES, our gradient estimation takes the sum of all candidate solutions’ displacements weighted by their corresponding standardized fitness (reward) scores [24]. We call this gradient-based evolution strategy Weighted Standard Rewards (WSR). Given the population size $n$, objective function $F$, mean reward $F(m)$, and standard deviation of rewards $\sigma'$, the gradient can be approximated by the following expression:

\begin{equation}
\nabla_\mu{}J(\mu) \approx \frac{1}{n}\sum_{i=0}^n \frac{F(\mu+\varepsilon)-F(m)}{\sigma'}*\varepsilon_i \label{eq9}
\end{equation}

Another method of gradient estimation involves using finite differences to numerically approximate the partial derivative of the loss with respect to every parameter. Interestingly, the finite differences method can easily be cast within the framework of an ES model, in which the total population size is set equal to the number of parameters plus one, and the ES sampling procedure is modified to apply finite differences to the parameters. In this modified sampling procedure, all candidate models would be near identical, except each model would have a single parameter—chosen based on a one-to-one mapping of parameters to candidate models—nudged with a small positive value $\sigma_{FD}=0.001$, while every other parameter is held constant. An additional candidate model is instantiated with its parameters set equal to the mean parameter vector of the population with no parameters changed. The fitness of this mean candidate model is used in place of $F(m)$ in \eqref{eq9}, and the finite difference value $\sigma_{FD}$ is used in place of the standard deviation $\sigma'$. The resulting modified gradient approximation expression is used to compute the gradient for finite differences. We explore the performance of finite differences as an additional training algorithm for ProtoNet, in order to compare ES to an alternative numerical optimization algorithm for few-shot learning.

\section{Experiment}

\subsection{Hardware}

Tensor Processing Units (TPUs) are a class of deep learning ASICs designed for compute-intensive neural network training. TPUs have been found to be, on average, 15 to 30 times faster than contempary GPUs and CPUs [25, 26]. TPUs and TPU pods are available through the Google Cloud Platofrm.

Each TPU device contains eight TPU cores, across which training can be parallelized and quickly synced. Each TPU core has either 8 gigabytes (in the case of TPUv2) or 16 gigabytes (in the case of TPUv3) of high bandwidth memory (HBM). Multiple TPU devices can be assembled together as TPU pods, in which constituent TPU devices are interconnected by high-speed mesh networks for fast communication. 

This design results in TPUs being geared toward highly parallel training tasks, specifically training with large batch sizes, or—as we demonstrate in this paper—population parallelism in evolutionary algorithms. The TPUs we used in our experiments were the v3-8 (single TPUv3 device containing 8 cores), and the v3-256 (TPUv3 pod containing 256 cores).

It is important to note that the TPU cluster size determines the number of shards across which the population in ES is partitioned. The number of partitions is equal to the number of TPU cores, such that a v3-8 with 8 cores will partition the population across 8 shards and a v3-256 with 256 cores will partition the population across 256 shards. The population size per core is a tuneable hyperparameter that we choose based on per-core memory constraints, and total population size is the per-core population size multiplied by the number of cores.

\subsection{Experiment}

\begin{table*}[!htbp]
\caption{Summary of Results}
\begin{center}
\begin{tabular}{ c c c c c c c c c }
\hline
% TPU & $\alpha$ & $\sigma$ & Channels & \begin{tabular}[x]{@{}c@{}c@{}}Pop.\\per\\core\end{tabular} & \begin{tabular}[x]{@{}c@{}}Total\\pop.\end{tabular} & Way & Shot & Accuracy \\
 TPU & $\alpha$ & $\sigma$ & Channels & Pop. per core & Total pop. & Way & Shot & Accuracy \\
\hline
 v3-8 & 1 & 0.01 & 16 & 64 & 512 & 20 & 1 & $0.964\pm0.013$ \\
 v3-256 & 10 & 0.1 & 16 & 64 & 16384 & 20 & 1 & $0.973\pm0.014$ \\
 ProtoNet & N/A & N/A & 64 & N/A & N/A & 20 & 1 & $0.987$ \\
\hline
 v3-8 & 1 & 0.01 & 32 & 32 & 256 & 10 & 5 & $0.990\pm0.005$ \\
 v3-256 & 10 & 0.1 & 16 & 64 & 16384 & 20 & 5 & $0.991\pm0.009$ \\
 ProtoNet & N/A & N/A & 64 & N/A & N/A & 20 & 5 & $0.996$ \\
\hline
%\multicolumn{8}{l}{$^{\mathrm{a}}$Sample of a Table footnote.}
\end{tabular}
\label{tabl1}
\end{center}
\end{table*}

The Omniglot dataset [27] is comprised of 1623 classes of handwritten characters, each of which containing 20 grayscale 105 × 105 images. The images were resized to 32 × 32 for our experiments, and the dataset was augmented using rotations of 90 degrees, 180 degrees, and 270 degrees. The original orientations were kept, so that the post-augmentation dataset is four times the size of the original dataset. The augmented dataset was randomly split into train, validation, and test sets with 4804, 1012, and 676 classes respectively. The results reported below are from the test set.

Each trial consisted of 200 epochs, each with 100 episodes of training and one episode of validation. Each trial had a 5-way test and either 1-shot or 5-shot setup during meta-testing. Test accuracies were determined by running the trained model on 200 episodes with classes randomly sampled from the test set classes, and the standard deviation of the each accuracy value is reported as well.

We varied variables such as TPU size and number of convolution channels to reflect the effect of hardware availability and desired model size on test performance. All other hyperparameters were chosen based on validation performance (highest validation accuracies) to find the ideal hyperparameter configurations for each given value of the variable under scrutiny. A set of high-performing hyperparameter configurations on the validation set for each given value of the variable under scrutiny was kept and re-evaluated on the test set. The resulting accuracies are reported throughout this section.

Our ES-ProtoNet used the same convolutional network architecture as Snell et al. [7]: a sequence of four identical convolution blocks, with each block comprised of a 3 × 3 convolution, batch normalization layer, ReLU nonlinearity, and 2x2 max-pooling layer in the same order that they are mentioned. The original ProtoNet used 64 channels for all of its convolution layers, but we experiment with 64, 32, and 16 channels across different trials. The rest of our experimental parameters were set up identically to the original ProtoNet paper, unless otherwise stated.

In Table 1, we summarize our results for the best performing models. In Table 2, we demonstrate how changes to the step size (alpha) and population standard deviation (sigma) parameters could cause test accuracy to deteriorate. In table 2 only, alpha and sigma values were varied while all other hyperparameters were held constant (rather than tuned using the validation set). The control variables are the Euclidean distance metric, the population size, the train way, and the number of convolutional filters or channel counts. The term “Pop. per core” will be used to refer to the population size per TPU core (shard/worker) in the tables, whereas the total population size is equal to the per-core population multiplied by the number of cores, and is referred to by the “Total pop.” term. The dependent variable is the accuracy and accompanying standard deviation.

\begin{table}[!htbp]  
\caption{Hyperparameter Optimization}
\begin{center}
\begin{tabular}{ c c c }
\hline
$\alpha$ & $\sigma$ & Accuracy \\
\hline
 0.1 & 0.001 & $0.62\pm0.028$ \\
 1 & 0.01 & $0.987\pm0.009$ \\
 10 & 0.1 & $0.971\pm0.011$ \\
 25 & 0.25 & Collapse \\
\hline
%\multicolumn{3}{p{150}}{$^{\mathrm{a}}$Effect of varying $\alpha$ and $\sigma$ on test accuracy, with all other variables held constant (TPU=v3-8, Channels=16, Pop. per core=128, Total pop.=1024, Way=10).}
\multicolumn{3}{p{150pt}}{Effect of varying $\alpha$ and $\sigma$ on test accuracy, with all other variables held constant (TPU=v3-8, Channels=16, Pop. per core=128, Total pop.=1024, Way=10).}
\end{tabular}
\label{tabl2}
\end{center}
\end{table}

It can be seen that different TPU models had different optimal alpha (step size) and sigma (population standard deviation) values. When alpha and sigma were too large, the test accuracy would collapse to random guessing. When they were too small, the model parameters could not escape a local minimum in the loss function.

We experimented with different numbers of train way (number of classes sampled per training episode) from 5, 10, and 20 on different TPU models with different alpha and sigma values as seen in Fig. 1. The control variables were the channel count and the test way. The population size per core was inversely proportional to the train way size. Refer to table 5 in the appendix for details.

A train way of 10 allowed for the highest performances for the v3-8, but a value of 20 allowed for the highest performance for the v3-256 for both alpha and sigma combinations. This is consistent with the conclusion that models are more robust when the test way is smaller than the train way, as proposed in Snell et al. [7].

We tested convolution with 16 channels, 32 channels, and 64 channels, with different TPUs as seen in Table 3. The constants were the alpha and sigma values. Because that model size depended heavily on channel count (parameter count in the convolution blocks grows as a quadratic function of channel count) while memory available per core was constant, we were able to use larger per-core population sizes when using smaller convolution channel counts. Conversely, we could use larger channel counts when using smaller per-core population sizes.

There was relatively little variation in accuracy performance based on the number of channels. 32 channels was optimal for v3-8, but 16 was optimal for v3-256.

Fig. 2 compares the training progress of v3-8 TPU models depending on the number of convolution channels. The train way was 10 for all three channel counts. The larger channel count led to faster convergence but had similar performance after 200 epochs.

The last set of experiments with our evolution strategy (WSR) was the one-shot tests, which can be found in Table 6 in the appendix.

\begin{figure}[htbp]
\centerline{\includegraphics[width=0.5\textwidth]{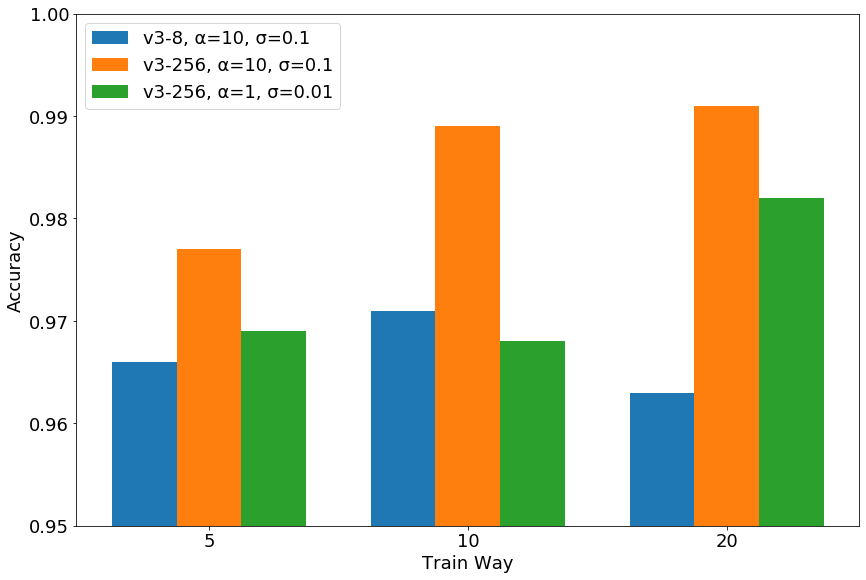}}
\caption{Modifying Train Way}
\label{fig1}
\end{figure}

\begin{table}[!htbp]
\caption{Modifying Number of Channels}
\begin{center}
\begin{tabular}{ c c c c c }
\hline
% TPU & $\alpha$ & $\sigma$ & Channels & \begin{tabular}[x]{@{}c@{}c@{}}Pop.\\per\\core\end{tabular} & \begin{tabular}[x]{@{}c@{}}Total\\pop.\end{tabular} & Way & Shot & Accuracy \\
 Channels & Pop. per core & Total pop. & Way & Accuracy \\
\hline
%\multicolumn{5}{c}{with TPU=v3-8, $\alpha=1$, $\sigma=0.01$} \\
\multicolumn{5}{c}{} \\
 64 & 16 & 128 & 5 & $0.978\pm0.011$ \\
 64 & 8 & 64 & 10 & $0.984\pm0.008$ \\
 32 & 64 & 512 & 5 & $0.98\pm0.009$ \\
 32 & 32 & 256 & 10 & $0.99\pm0.005$ \\
 32 & 16 & 128 & 20 & $0.987\pm0.009$ \\
 16 & 128 & 1024 & 10 & $0.987\pm0.009$ \\
\multicolumn{5}{c}{(with TPU=v3-8, $\alpha=1$, $\sigma=0.01$)} \\
\hline
%\multicolumn{5}{c}{with TPU=v3-256, $\alpha=10$, $\sigma=0.1$} \\
\multicolumn{5}{c}{} \\
 32 & 32 & 8192 & 10 & $0.989\pm0.008$ \\
 32 & 16 & 4096 & 20 & $0.988\pm0.01$ \\
 16 & 64 & 16384 & 20 & $0.991\pm0.009$ \\
 16 & 128 & 32768 & 10 & $0.989\pm0.008$ \\
\multicolumn{5}{c}{(with TPU=v3-256, $\alpha=10$, $\sigma=0.1$)} \\
\hline
%\multicolumn{8}{l}{$^{\mathrm{a}}$Sample of a Table footnote.}
\end{tabular}
\label{tabl3}
\end{center}
\end{table}

\begin{figure}[htbp]
\centerline{\includegraphics[width=0.5\textwidth]{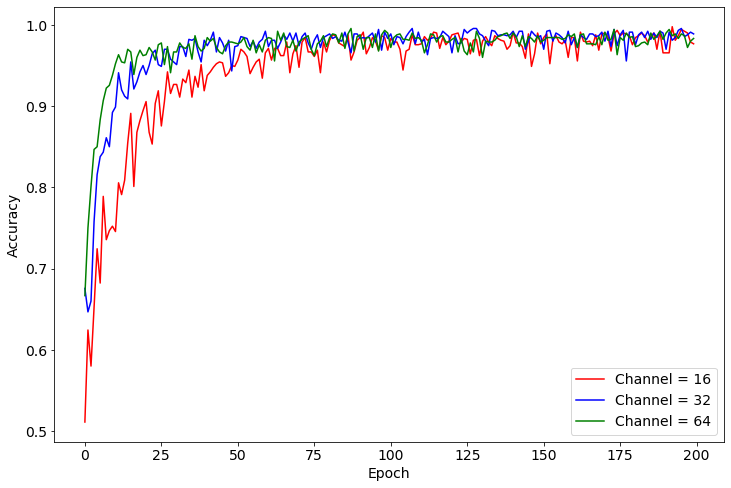}}
\caption{Convergence Speed Based on Number of Channels}
\label{fig2}
\end{figure}

In the finite differences strategy experiments, it was found that the Euclidean distance metric achieved substantially lower accuracy compared to cosine similarity. The independent variables were the learning rate, population size, and train way.  In Table 4 we display the highest performing instances of the finite differences trials. 

Because that the finite differences algorithm significantly underperformed compared to evolution strategies in 5-shot classification, 1-shot trials were not attempted.

Lastly, different learning rates of the finite differences algorithm were investigated. The results in Fig. 3 are from cosine similarity calculations because Euclidean distance was found to significantly underperform. In addition, different population sizes and train way configurations were tried, as shown in Table 7 in the appendix. The optimal learning rate for v3-8 was 5×10-03, and 5×10-04 for v3-256.

We observed a clear disparity in performance between finite differences and evolution strategies on five-shot ProtoNet classification. Although finite differences ensures that the search directions of all candidate models are orthogonal to each other in parameter space, the same property confines the candidate models to a limited number of directions. In contrast, evolution strategies allows for a potentially infinite number of search directions. However, based on our experimental results, we believe that the infinite possible search directions evaluated allows for more accurate gradient estimations than finite differences.

\begin{table}[!htbp]
\caption{Finite Differences 5-Shot Results}
\begin{center}
\begin{tabular}{ c c c c c }
\hline
 TPU & Distance & Learning Rate & Way & Accuracy \\
\hline
 v3-8 & Cosine & 5E-03 & 10 & $0.939\pm0.012$ \\
 v3-8 & Cosine & 1E-02 & 20 & $0.944\pm0.014$ \\
 v3-256 & Cosine & 5E-04 & 20 & $0.956\pm0.014$ \\
 v3-8 & Euclidean & 1E-03 & 10 & $0.853\pm0.026$ \\
 v3-256 & Euclidean & 1E-03 & 10 & $0.855\pm0.028$ \\
\hline
\end{tabular}
\label{tabl4}
\end{center}
\end{table}

\begin{figure}[htbp]
\centerline{\includegraphics[width=0.5\textwidth,height=6cm, keepaspectratio]{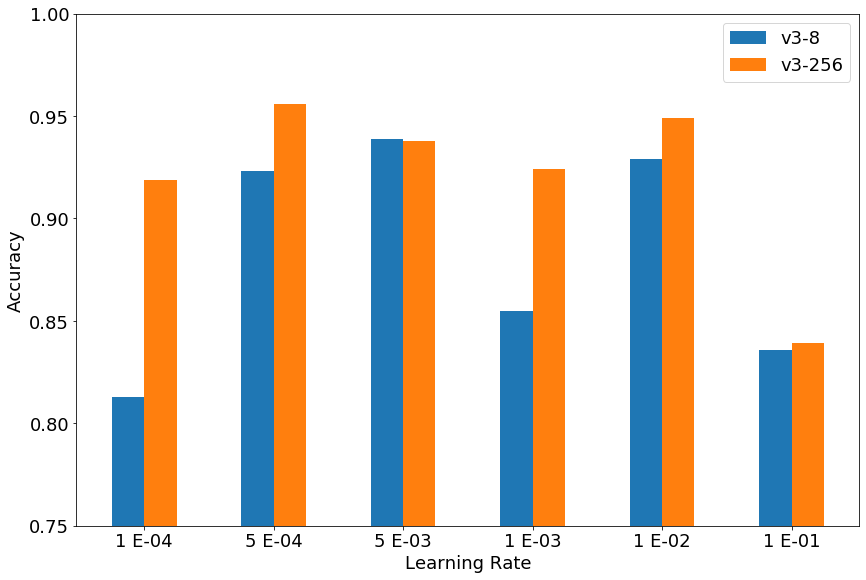}}
\caption{Modifying Finite Differences (Cosine Similarity Metric) Learning Rate}
\label{fig3}
\end{figure}

\section{Conclusion}

We experimentally validated the viability of evolution strategies as an alternative to backpropagation in supervised few-shot learning. We successfully used ES to train a prototypical network to perform competitively compared to a backpropagation-trained counterpart.

We observed an almost negligible difference in performance (~0.5\text{\%} accuracy drop) from backpropagation-trained Protonet, which we attribute to the natural inaccuracy of gradients estimated by evolution strategies in comparison to exact methods like backpropagation. We were able to achieve competitive results with many different hyperparameter configurations (including different model sizes, numbers of workers/cores, population sizes, step sizes, and population sampling standard deviation values), demonstrating the stability of ES-ProtoNet convergence across a broad variety of training conditions.

We observed especially interesting results with some of the computationally cheaper training configurations we used, particularly one which involved training a ProtoNet with 64 channels using a total population size of 64. In this particular run, the memory cost of the ES population matrix was approximately forty times smaller than the memory cost of the dominant jacobian matrix in the equivalent analytical gradient computation. With theoretically as little as 3\text{\%} of the memory cost, ES-ProtoNet was able to achieve 98.4\text{\%} accuracy on 5-shot Omniglot classification (as shown in Table 3), within 1.3\text{\%} of the original ProtoNet at 99.6\text{\%}.

By conceding a marginal amount of performance, it is possible to achieve meta-learning that not only has constant memory complexity with respect to task length, but is also easily parallelizable along the population axis (potentially distributing the computational cost better than any other parallelization method). Furthermore, the performance difference can be almost completely rectified either by using a large population size, or by accumulating gradients over multiple population steps to achieve a large effective population size.

Such a result is promising, as it demonstrates the possibility of eliminating the backward step in meta-learning while only storing the population matrix, which can be made smaller than the jacobian matrix necessary for forward-mode differentiation. Using ES, it is possible to perform meta-learning on arbitrarily complex tasks without bearing the full memory complexity of forward-mode differentiation, while also distributing most of the computational cost over large TPU clusters. We hope that these results will encourage future work to use evolution-based algorithms to scale up meta-learning to more complex and impactful task domains.

\section*{Acknowledgment}

Our research was supported with Cloud TPUs from Google’s TensorFlow Research Cloud (TFRC).

\section*{Appendix}

In Table 5, an extended collection of results from modifying the train way is shown. The population size was varied inversely with the train way to maintain relatively constant memory cost per core.

\begin{table}[htbp]
\caption{Modifying Train Way}
\begin{center}
\begin{tabular}{ c c c c c c }
\hline
 TPU & $\alpha$ & $\sigma$ & Total pop. & Way & Accuracy \\
\hline
 v3-8 & 10 & 0.1 & 512 & 20 & $0.963\pm0.012$ \\
 v3-8 & 10 & 0.1 & 1024 & 10 & $0.971\pm0.011$ \\
 v3-8 & 10 & 0.1 & 2048 & 5 & $0.966\pm0.011$ \\
 v3-256 & 10 & 0.1 & 16384 & 20 & $0.991\pm0.009$ \\
 v3-256 & 10 & 0.1 & 32768 & 10 & $0.989\pm0.008$ \\
 v3-256 & 10 & 0.1 & 65536 & 5 & $0.977\pm0.010$ \\
 v3-256 & 1 & 0.01 & 16384 & 20 & $0.982\pm0.010$ \\
 v3-256 & 1 & 0.01 & 32768 & 10 & $0.968\pm0.011$ \\
\hline
\multicolumn{6}{c}{All of the above are using 16-channel convolutions.} \\
\end{tabular}
\label{tabl5}
\end{center}
\end{table}

In Table 6, an extended collection of results from the one-shot learning experiments is presented.

\begin{table}[htbp]
\caption{One-Shot Results}
\begin{center}
\begin{tabular}{ c c c c c c }
\hline
 $\alpha$ & $\sigma$ & Channels & Total pop. & Way & Accuracy \\
\hline
\multicolumn{6}{c}{} \\
 1 & 0.01 & 16 & 1024 & 10 & $0.954\pm0.016$ \\
 1 & 0.01 & 16 & 512 & 20 & $0.964\pm0.013$ \\
 1 & 0.01 & 32 & 256 & 10 & $0.958\pm0.014$ \\
 1 & 0.01 & 32 & 512 & 20 & $0.960\pm0.014$ \\
 10 & 0.1 & 16 & 1024 & 10 & $0.902\pm0.023$ \\
 10 & 0.1 & 16 & 512 & 20 & $0.876\pm0.024$ \\
\multicolumn{6}{c}{(with TPU=v3-8)} \\
\hline
\multicolumn{6}{c}{} \\
 10 & 0.1 & 32 & 4096 & 20 & $0.964\pm0.015$ \\
 10 & 0.1 & 32 & 8192 & 10 & $0.961\pm0.014$ \\
 10 & 0.1 & 16 & 16384 & 20 & $0.973\pm0.014$ \\
\multicolumn{6}{c}{(with TPU=v3-256)} \\
\hline
%\multicolumn{6}{c}{} \\
\end{tabular}
\label{tabl6}
\end{center}
\end{table}

\begin{figure}[htbp]
\centerline{\includegraphics[width=0.5\textwidth]{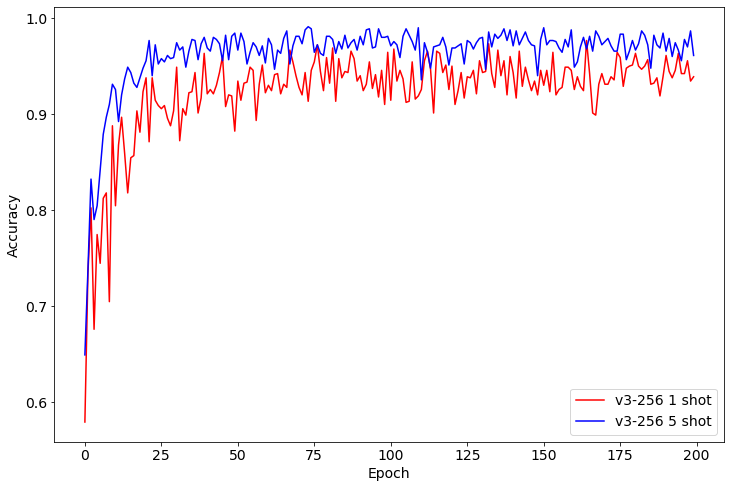}}
\caption{One-Shot versus Five-Shot Learning for v3-256 TPU}
\label{fig}
\end{figure}

In Fig.  4, the one-shot versus five-shot accuracy curves are displayed. They reached convergence at roughly the same time, but naturally did not obtain the same final accuracies.

In Table 7, the extended results of the finite differences tests using the cosine similarity metric are presented. The tests using the Euclidean distance metric are not reported as five independent trials could not obtain above an 86\text{\%} accuracy, and thus different learning rates were not tested.

\begin{table}[htbp]
\caption{Finite Differences: Learning Rate versus Test Accuracy}
\begin{center}
\begin{tabular}{ c c c c c c c }
\hline
 TPU & Total pop. & Way & $5E-04$ & $5E-03$ & $1E-03$ & $1E-02$ \\
\hline
 v3-8 & 1024 & 10 & 0.923 & 0.939 & 0.855 & 0.929 \\
 v3-256 & 16384 & 20 & 0.956 & 0.938 & 0.924 & 0.949 \\
 v2-512 & 32768 & 10 & 0.943 & 0.935 & 0.949 & 0.946 \\
 v3-8 & 512 & 20 & 0.925 & 0.94 & 0.821 & 0.944 \\
\hline
\multicolumn{7}{p{250pt}}{Effect of different learning rates on test accuracy in finite differences ProtoNet. All of the above are using cosine similarity.} \\
\end{tabular}
\label{tabl7}
\end{center}
\end{table}

In Fig. 5, the accuracy curve of our evolution strategy (denoted as WSR) was compared to finite differences.

\begin{figure}[htbp]
\centerline{\includegraphics[width=0.5\textwidth]{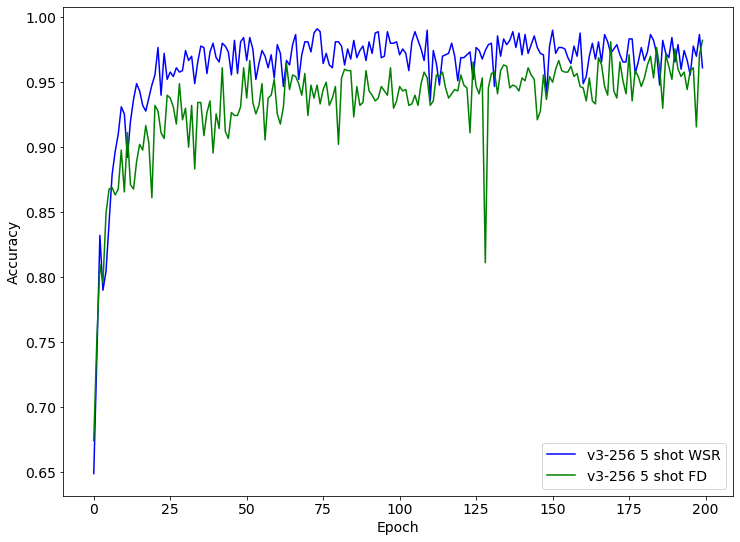}}
\caption{Evolution Strategy versus Finite Differences}
\label{fig6}
\end{figure}

\end{document}